\documentclass[letterpaper]{article} % DO NOT CHANGE THIS
\usepackage{aaai23}  % DO NOT CHANGE THIS
\usepackage{times}  % DO NOT CHANGE THIS
\usepackage{helvet}  % DO NOT CHANGE THIS
\usepackage{courier}  % DO NOT CHANGE THIS
\usepackage[hyphens]{url}  % DO NOT CHANGE THIS
\usepackage{graphicx} % DO NOT CHANGE THIS
\usepackage{natbib}  % DO NOT CHANGE THIS AND DO NOT ADD ANY OPTIONS TO IT
\usepackage{caption} % DO NOT CHANGE THIS AND DO NOT ADD ANY OPTIONS TO IT
\usepackage{algorithm}
\usepackage[ruled,linesnumbered,algo2e]{algorithm2e}
\usepackage{todonotes}
\usepackage{amsmath}
\usepackage{mathtools}
\usepackage{booktabs}
\usepackage{amsfonts}
\usepackage{pifont}
\usepackage{wasysym}
\usepackage{multirow}
\usepackage{chemfig}

\newcommand{\set}[1]{\mathcal{#1}}
\usepackage{newfloat}
\usepackage{listings}
\DeclareCaptionStyle{ruled}{labelfont=normalfont,labelsep=colon,strut=off} % DO NOT CHANGE THIS
\floatstyle{ruled}
\newfloat{listing}{tb}{lst}{}
\floatname{listing}{Listing}
\title{Retrosynthesis Prediction with Local Template Retrieval}
\author{
    Shufang Xie\textsuperscript{\rm 1}, Rui Yan\textsuperscript{\rm 1}\thanks{Corresponding author: Rui Yan (ruiyan@ruc.edu.cn).}, Junliang Guo\textsuperscript{\rm 2}, Yingce Xia\textsuperscript{\rm 3}, Lijun Wu\textsuperscript{\rm 3}, Tao Qin\textsuperscript{\rm 3} \\
}
\affiliations{
    \textsuperscript{\rm 1}Beijing Key Laboratory of Big Data Management and Analysis Methods,\\ Gaoling School of Artificial Intelligence (GSAI),\, Renmin University of China \\
    \textsuperscript{\rm 2}Microsoft Reserarch Aisa\\
    \textsuperscript{\rm 3}Microsoft Reserarch AI4Science \\
    \{shufangxie,ruiyan\}@ruc.edu.cn, \{junliangguo,yingce.xia,lijuwu,taoqin\}@microsoft.com
    
}

\usepackage{bibentry}
\begin{document}

\maketitle

\begin{abstract}
Retrosynthesis, which predicts the reactants of a given target molecule, is an essential task for drug discovery. 
In recent years, the machine learing based retrosynthesis methods have achieved promising results.
In this work, we introduce RetroKNN, a local reaction template retrieval method to further boost the performance of template-based systems with non-parametric retrieval. We first build an atom-template store and a bond-template store that contain the local templates in the training data, then retrieve from these templates with a k-nearest-neighbor (KNN) search during inference. The retrieved templates are combined with neural network predictions as the final output.
Furthermore, we propose a lightweight adapter to adjust the weights when combing neural network and KNN predictions conditioned on the hidden representation and the retrieved templates.
We conduct comprehensive experiments on two widely used benchmarks, the USPTO-50K and USPTO-MIT.
Especially for the top-1 accuracy, we improved 7.1\% on the USPTO-50K dataset and 12.0\% on the USPTO-MIT dataset.
These results demonstrate the effectiveness of our method.
%On the USPTO-50K dataset, we improved the top-1 accuracy from 53.4 to 57.2 and from 54.1 to 60.6 on USPTO-MIT. 
%Our code is available at xxxx.
\end{abstract}

\section{Introduction}
Retrosynthesis, which predicts the reactants for a given product molecule, is a fundamental task for drug discovery. The conventional methods heavily rely on the expertise and heuristics of chemists~\cite{corey1991logic}. Recently, machine learning based approaches have been proposed to assist chemists and have shown promising results~\cite{dong_deep_2021}.
The typical approaches includes the template-free methods that predict the reactants directly and the template-based methods that first predict reaction templates and then obtain reactants based on templates.
For these different approaches, a shared research challenge is effectively modeling this task's particular property.
 
\begin{figure}[!htbp]
    \centering
    \includegraphics[width=0.95\linewidth]{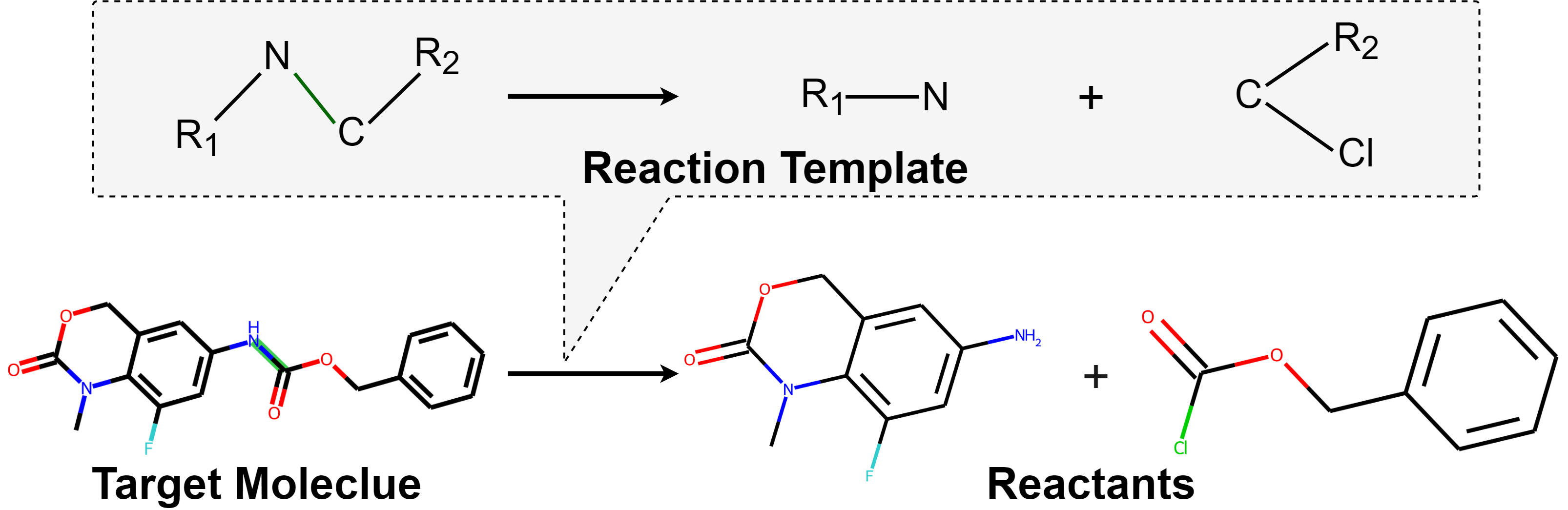}
    \caption{Illustration of retrosynthesis that takes the target molecule on the left side and predicts two reactants on the right side. Inside the callout is its reaction template that breaks the carbon-nitrogen bond into two parts.}
    \label{fig:example}
\end{figure}

As shown in Figure~\ref{fig:example}, a key property of a chemical reaction is that it is strongly related to modifying the local structure of the target molecule, such as replacing a functional group or breaking a bond. 
Therefore, much recent research focuses on better modeling the local structure of molecules~\cite{chen_deep_2021,somnath_learning_2021}.
Despite their promising results, we notice that it is still challenging to learn all reaction patterns only with neural networks, especially for the rare templates. 

Therefore, we introduce a non-parametric retrieval-based method to provide concrete guidance in prediction. Specifically, we use a local template retrieval method, the k-nearest-neighbor (KNN) method, to provide additional predictions to improve the prediction accuracy. 
Following LocalRetro~\cite{chen_deep_2021}, We first take a trained graph-neural network~(GNN) for the retrosynthesis task and offline build an atom-template and a bond-template store that contain reaction templates~(Section~\ref{sec:prelim}).
During this store construction phase, we iterate all target molecules in the training data and add the templates of each atom and each bond to the corresponding store. The templates are indexed by the hidden representations extracted by the GNN. During inference, for a given new target molecule, we first use the original GNN to extract the hidden representations as well as the original GNN predicted templates. Then, we use the hidden representations to search the two stores to retrieve local templates similar to the query.
The GNN predicted templates and the KNN retrieved templates are merged with different weights to build the final output.

Combining the GNN and KNN predictions is one key design factor in the above processes. The conventional way is to use fixed parameters to aggregate these predictions for all reactions, which may be sub-optimal and hurt the model's generalization~\cite{zhen2021ada}. Because each prediction may have a different confidence level, it would be beneficial to assign the weights adaptively for each reaction across different instances (Section~\ref{sec:case}). 
Therefore, we employ a lightweight adapter to predict these values conditioned on the GNN representations and the retrieved results.
The adapter network has a simple structure and is trained with a few samples.
Although the adapter has a little extra cost, it can help improve the model performance effectively.

To sum up, our contribution is two fold:
\begin{itemize}
    \item We propose RetroKNN, a novel method to improve the retrosynthesis prediction performance with local template retrieval by the non-parametric KNN method.
    \item We propose a lightweight meta-network to adaptively control the weights when combining the GNN and KNN predictions.
\end{itemize}

We conduct experiments on two widely used benchmarks: the USPTO-50K and USPTO-MIT. These datasets contain organic reactions extracted from the United States Patent and Trademark Office (USPTO) literature. On the USPTO-50K dataset, we improve the top-1 accuracy from 53.4 points to 57.2 points (7.1\% relative gain) and achieved new state-of-the-art. Meanwhile, on USPTO-MIT, we improve the top-1 accuracy from 54.1 points to 60.6 points (12.0\% relative gain).
Moreover, our method shows promising results on the zero-shot and few-shot datasets, which are challenging settings for conventional template-based methods yet essential for this research field.
These results demonstrate the effectiveness of our method.

\section{Method}
\subsection{Preliminaries}
\label{sec:prelim}
We denote a molecule as a graph $G(\set{V}, \set{E})$ where the $\set{V}$ is the node set and the $\set{E}$ is the bond set.
Given a target molecule $M$ as input, the retrosynthesis prediction task is to generate molecules set $\set{R}$ that are reactants of $M$.
Instead of directly predicting $\set{R}$, we follow LocalRetro~\cite{chen_deep_2021} that predict a local reaction template $t$ at reaction center $c$ and apply $(t, c)$ to molecule $M$.
More specifically, the $t$ is classified into two types: \textbf{a}tom-template $t \in \set{T}_{a}$ and \textbf{b}ond-template $t\in\set{T}_{b}$, depending whether $c$ is an atom or a bond.

%The template $t$ is selected from an \textbf{a}tom template set $\set{T}_{a}$ when $c$ is an atom or a \textbf{b}ond template set $\set{T}_{b}$ when $c$ is a bond.

We also assume that there are a training set $\set{D}_{\text{train}}$, a validation set $\set{D}_{\text{val}}$, and a test set $\set{D}_{\text{test}}$ available.
Each data split contains the target and corresponding reactants, which is formulated as $\set{D} = \{ (M_i, t_i, c_i, \set{R}_i)\}_{i=1}^{|\set{D}|}$ where $c_i$ is the reaction center of $M_i$ to apply the template $t_i$ and  $|\set{D}|$ is the data size of $\set{D}$.

Meanwhile, we assume a GNN model trained on $\set{D}_{\text{train}}$ exist. Without loss of generality, we split the GNN into two parts: a feature extractor $\mathbf{f}$ and a prediction head $\mathbf{h}$.
The feature extractor $\mathbf{f}$ takes a molecule graph $G(\set{V}, \set{E})$ as input and output hidden representations $h_v$ for each node $v \in \set{V}$ and $h_e$ for each edge $e \in \set{E}$.
The $h_v$ and $h_e$ are processed by prediction head $\mathbf{h}$ to predict the probability distribution over the template set $\set{T}_a$ and $\set{T}_b$, respectively.

%In this section, we will introduce the details of our method.
%The overview of our system is available in Figure~\ref{fig:system}.

% \definecolor{GNN}{HTML}{666666}
% \definecolor{knn}{HTML}{000099}
% \definecolor{adapter}{HTML}{006600}
% \definecolor{merge}{HTML}{994C00}
% \definecolor{ret}{HTML}{F19C99}

\begin{figure*}[!htbp]
    \centering
    \includegraphics[width=\linewidth]{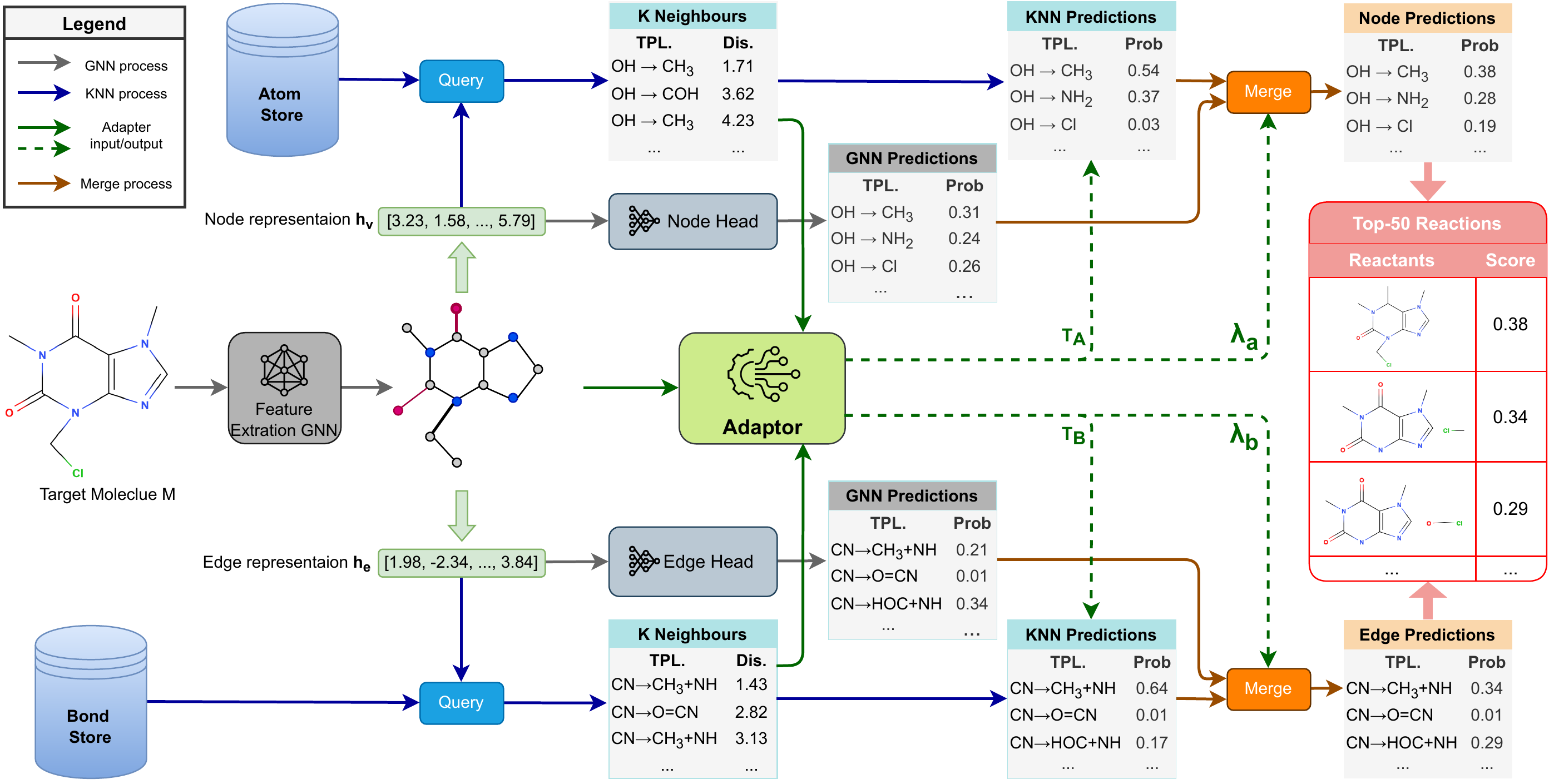}
    %\caption{The illustration of RetroKNN for a target molecule in middle left. The top and bottom half show the example of one atom and bond retrieval. The \textcolor{knn}{blue lines} denote the KNN process, the solid/dash \textcolor{adapter}{green lines} denote the input/output of adapter, and the \textcolor{merge}{brown lines} denote merging the predictions. The \textcolor{ret}{pink table} denotes the final output from all predictions.}
    %\caption{The illustration of RetroKNN for a target molecule in the middle left. The top and bottom half show the examples of one atom and bond retrieval. The \textcolor{GNN}{gray}, \textcolor{knn}{blue}, \textcolor{adapter}{green}, and \textcolor{merge}{brown} lines denote the  \textcolor{GNN}{GNN prediction}, \textcolor{knn}{KNN prediction}, \textcolor{adapter}{adapter input/output}, and \textcolor{merge}{ merge process}, respectively. The \textcolor{ret}{pink table} denotes the final output from all predictions.}
    \caption{The illustration of RetroKNN for a target molecule in the middle left. The top and bottom half show the examples of one atom and bond retrieval. The gray, blue, green, and brown lines denote the GNN prediction, KNN prediction, adapter input/output, and merge process, respectively. The pink table denotes the final output from all predictions.}
    \label{fig:system}
\end{figure*}

\subsection{Store Construction}

\begin{algorithm2e}[tb]
\caption{store construction algorithm}
\label{alg:store_create}
\KwIn{Training data $\set{D}_{\text{train}}$.}
\KwIn{Feature extractor $\mathbf{f}$.}
\KwOut{Atom store $\set{S}_A$ and bond store $\set{S}_B$.}
\textbf{Let} $\set{S}_A \coloneqq \emptyset, \set{S}_B\coloneqq\emptyset$ \tcp*[l]{Initialize.}
\For{$(M, t, c, \set{R}) \in \set{D}_{\text{train}}$}{
    \textbf{Let} $\set{V}$ denotes the node set of $M$\;
    \textbf{Let} $\set{E}$ denotes the edge set of $M$\;

    \For{$v \in \set{V}$ \tcp*[l]{Loop each node.}}{ 
        \textbf{Let} $h_v \coloneqq \mathbf{f}(v | M)$\;
        \eIf{$v == c$}{
            \textbf{Let} $\set{S}_A \coloneqq \set{S}_A \cup \{ (h_v, t) \} $\;
        }{
            \textbf{Let} $\set{S}_A \coloneqq \set{S}_A \cup \{ (h_v, \mathbf{0}) \}$\;
        }
    }
    
    \For{$e \in \set{E}$\tcp*[l]{Loop each edge.}}{
        \textbf{Let} $h_e \coloneqq \mathbf{f}(e | M)$\;
        \eIf{$e == c$}{
            \textbf{Let} $\set{S}_B \coloneqq \set{S}_B \cup \{ (h_e, t) \}$\;
        }{
         \textbf{Let} $\set{S}_B \coloneqq \set{S}_B \cup \{ (h_e, \mathbf{0}) \}$\;
        }
    }
}
\textbf{return} $\set{S}_A$, $\set{S}_B$
\end{algorithm2e}

Our method uses two data store $\set{S}_A$ and $\set{S}_B$ that contain the information of atoms and bonds.
Both of the store are constructed offline before inference.
Inside the store are key-value pairs that are computed from $\set{D}_{\text{train}}$ and the construction procedure details are in Algorithm~\ref{alg:store_create}.

In this algorithm, the first step is to initialize the atom store $\set{S}_A$ and bond store $\set{S}_B$ as an empty set.
Next, for each reaction in the training data $\set{D}_{\text{train}}$, we iterate all nodes $v \in \set{V}$ and all edges $e \in \set{E}$ of the target molecule $M$ in line 5 to 13 and line 14 to 22, respectively.
For each node $v$, if it is the reaction center, we add template $t$ that indexed by the hidden representation $h_v$ to the $\set{S}_A$.
Otherwise, we add a special token $\mathbf{0}$ to indicate that template is not applied here.
Similarly, for each edge $e$, we add either $(h_e, t)$ or $(h_e, \mathbf{0})$ to the bond store $\set{S}_B$.
Finally, we get the atom store $\set{S}_A$ and the bond store $\set{S}_B$ used during inference.
%Because these stores are built from the training data, there is no risk of a data leak.

\subsection{Inference Method}
The overview of inference procedure is available in Figure~\ref{fig:system}.
At inference time, given a new target molecule $M$, we first compute the hidden representation $h_v, h_e$ and template probability $P_{\text{GNN}}(t_a | M, a), P_{\text{GNN}}(t_b|M, b)$ for each atom $a$ and bond $b$, respectively\footnote{Whenever possible, we omit the subscript of node and edge id to simplify the notations.}.
Next, we retrieve the templates for each node and edge, which can be written as

{\footnotesize
\begin{align}
    P_{\text{KNN}}(t_a|M, a) \propto \sum_{ (h_i, t_i) \in \set{N}_a} \mathbb{I}_{t_a = t_i} \exp \left( \frac{-d(h_a, h_i) }{T_A}  \right), \label{eq:pta} \\
    P_{\text{KNN}}(t_b|M, b) \propto \sum_{ (h_i, t_i) \in \set{N}_b} \mathbb{I}_{t_b = t_i} \exp \left( \frac{-d(h_b, h_i) }{T_B}  \right). \label{eq:ptb}
\end{align}
}
In Equations~(\ref{eq:pta}, \ref{eq:ptb}), the $\set{N}_a, \set{N}_b$ are candidates sets that retrieved from $\set{S}_A, \set{S}_B$, the $\mathbb{I}$ is the indicator function that only outputs 1 when the condition (i.e., $t_a = t_i$ or $t_b = t_i$) is satisfied, and the $T_A, T_B$ are the softmax temperate.
Meanwhile, the $d(\cdot, \cdot)$ is the distance function to measure the similarity between $h_i$ with $h_v$ or $h_e$.
In another words, the $P_{\text{KNN}}(t_a|M, a)$ is proportional to the sum of the weights of the neighbours whose template is $t_a$.

Finally, we combine the GNN output and KNN output with interpolation factors $\lambda$, which is
{\footnotesize
\begin{align}
    P(t_a|M,a) &= \lambda_a P_{\text{GNN}}(t_a | M,a)\! +\! (1\! -\! \lambda_a)P_{\text{KNN}}(t_a|M,a), \label{eq:tam} \\
    P(t_b|M,b) &= \lambda_b P_{\text{GNN}}(t_b | M,b) + (1\!-\!\lambda_b)P_{\text{KNN}}(t_b|M,b). \label{eq:tbm}
\end{align}
}

In the Equation~\eqref{eq:pta}-\eqref{eq:tbm}, the temperature $T_A, T_B \in \mathbb{R}^+$ and the interpolation factors $\lambda_a, \lambda_b \in [0, 1]$ are predicted by the adaptor network and details are introduced in Section~\ref{sec:adapter}.

In Figure~\ref{fig:system}, we only illustrate one node and one bond retrieval as examples, but in practice, we conduct such a process for all atoms and bonds.  Following LocalRetro~\cite{chen_deep_2021}, after we get the $P(t_a|M,a)$ and $P(t_b|M,b)$ for each atom $a$ and bond $b$, we will rank all non-zero predictions by their probability.
The atom template and bonds templates are ranked together, and the top 50 predictions are our system's final output.

\subsection{Adaptor Network}
\label{sec:adapter}
To adaptively choose the $T_A, T_B, \lambda_a$, and $\lambda_b$ for each atom and bond, we design a lightweight network to predict these values.
The input to adapter are hidden representation $h_v, h_e$ from GNN side and distance list $d(h_v, h_i), d(h_e, h_i)$ from the KNN side.
%These features capture the local information for the GNN and KNN prediction.

We use a one-layer GNN followed by a few fully connected (FC) layers for the network architecture.
We use the the graph isomorphism network (GIN) with edge features ~\cite{hu2019strategies} layer to capture both node feature $h_v$ and edge feature $h_e$, which is formulated as:
\begin{equation}
    h_v^{(g)} = W_{\text{vg}}((1+\epsilon)h_v + \sum_{e\in\set{E}(v)} \text{ReLU}(h_v + h_e)) + b_{\text{vg}},
\end{equation}
where the $h_v^{(g)}$ is the output, $\epsilon$ and $W$ are learnable parameters of GIN, and the $\set{E}(v)$ is the set of edges around $v$.
Meanwhile, we use the FC layer to project the KNN distances to extract the features that can be formulated as
\begin{align}
    h_v^{(k)} = W_{\text{vk}}(\{d(h_v, h_i)\}_{i=1}^{K}) + b_{\text{vk}}, \\
    h_e^{(k)} = W_{\text{ek}}(\{d(h_e, h_i)\}_{i=1}^{K}) + b_{\text{ek}},
\end{align}
where the brackets $\{\cdot\}_{i=1}^{K}$ means building a K-dimensional vector.
Finally, the feature from GNN and KNN are combined to a mixed representation, which are
\begin{align}
    h_v^{(o)} = \text{ReLU}( W_{\text{vo}} \text{ReLU}(h_v^{(g)} \Vert h_v^{(k)}) + b_{\text{vo}}), \\
    h_e^{(o)} = \text{ReLU}( W_{\text{eo}} \text{ReLU}(h_{es}^{(g)} \Vert h_{et}^{(g)} \Vert h_e^{(k)}) + b_{\text{eo}}),
\end{align}
where the $\Vert$ denotes tensor concatenation and {\it es} and {\it et} are start and end node of edge $e$.

The $T_A, \lambda_a$ are predicted by  $h_v^{(o)}$ and the $T_B, \lambda_b$ are predicated by $h_e^{(o)}$ by another FC layer.
We also use sigmoid function $\sigma$ to guarantee the $\lambda_a, \lambda_b \in (0, 1)$ and clamp the $T_A, T_B$ into range $[1, 100]$.
Formally, we have
{\footnotesize
\begin{align}
    T_A &= \max(1, \min(100, W_{\text{ta}} h_v^{(o)} + b_{\text{ta}})), \\
    \lambda_a &= \sigma (W_{\text{la}} h_v^{(o)} + b_{\text{la}}), \\
    T_B &= \max(1, \min(100, W_{\text{tb}} h_e^{(o)} + b_{\text{tb}}, 1, 100)), \\
    \lambda_b &= \sigma(W_{\text{lb}} h_e^{(o)} + b_{\text{lb}}).
\end{align}
}

Because all the formulas used here are differentiable, we optimize the adapter parameters $W$ with gradient decent to minimize the template classification loss
{\footnotesize
\begin{align}
    \set{L}_M = &- \frac{1}{|\set{V}|} \sum_{a \in \set{V}} \log P(\hat{t}_a|M, a) \nonumber \\
        &- \frac{1}{|\set{E}|}\sum_{b \in \set{E}} \log P(\hat{t}_b|M, b),
\end{align}
}
for each target molecule M with node set $\set{V}$ and edge set $\set{E}$. The $ P(\hat{t}_a|M), P(\hat{t}_b|M)$ are computed by Equation~\eqref{eq:tam} and Equation~\eqref{eq:tbm}. The $\hat{t}_a, \hat{t}_b$ are the ground truth template.

\section{Experiments}
\subsection{Experimental Settings}

\paragraph{Data.}
Our experiments are based on the chemical reactions extracted from the United States Patent and Trademark Office (USPTO) literature.
We use two versions of the USPTO benchmark: the USPTO-50K~\cite{coley_computer-assisted_2017} and USPTO-MIT~\cite{jin2017predicting}.
The USPTO-50K contains 50k chemical reactions, split into 40k/5k/5k reactions as training, validation, and test, respectively.
Meanwhile, the USPTO-MIT consists of about 479k reactions, and the split is 409k/40k/30k.
All the partitions are the same as in previous works~\cite{coley_computer-assisted_2017,jin2017predicting} to make fair comparisons.
We also use the preprocess scripts by~\citet{chen_deep_2021} to extract the reaction templates from these reactions, which leads to 658 and 20,221 reaction templates in USPTO-50K and USPTO-MIT. 

\paragraph{Implementation details.}
We follow the same model configuration as LocalRetro~\cite{chen_deep_2021} to build the backbone GNN model.
The feature extractor $\mathbf{f}$ is a 6-layer \texttt{MPNN}~\cite{gilmer2017neural} followed by a single \texttt{GRA} layer~\cite{chen_deep_2021} with 8 heads.
We use the hidden dimension 320 and dropout 0.2.
The atoms' and bonds' input feature is extracted by DGL-LifeSci~\cite{dgllife}.% by \texttt{WeaveAtomFeaturizer} and \texttt{CanonicalBondFeaturizer}.
The prediction head $h$ consists two dense layers with \text{ReLU} activation.
The backbone model is optimized by Adam optimizer with a learning rate of 0.001 for 50 epochs. 
We also early stop the training when there is no improvement in the validation loss for five epochs.
The configurations for backbone are all same as~\citet{chen_deep_2021}.

The implementation of KNN is based on the faiss~\cite{johnson2019billion} library with \texttt{IndexIVFPQ} index for fast embedding searching, and the K of KNN is set to 32.
For the adapter network, we use the same hidden dimension as the backbone GNN.
The adapter is also trained with Adam optimizer with a learning rate of 0.001.
Considering the data size difference, we train the adapter for ten epochs and two epochs on the validation set of the USPTO-50K and USPTO-MIT datasets, respectively.
The adapter with the best validation loss is used for test.

\paragraph{Evaluation and baselines}
Following previous works, our system will predict top-50 results for each target molecule and report the top-K accuracy where K=1,3,5,10, and 50 by the script from~\citet{chen_deep_2021}. We also use representative  baseline systems in recent years, include:

%\begin{itemize}
\noindent $\bullet$ Template-based methods: retrosim~\cite{coley_computer-assisted_2017}, neuralsym~\cite{segler_neural-symbolic_2017}, GLN~\cite{dai_retrosynthesis_2020}, Hopfield~\cite{seidl_modern_2021}, and LocalRetro~\cite{chen_deep_2021};

\noindent $\bullet$  Semi-template based methods: G2Gs~\cite{shi_graph_2021}, RetroXpert~\cite{yan2020retroxpert}, and GraphRtro~\cite{somnath_learning_2021};

\noindent $\bullet$ Tempate-free methods: Transformer~\cite{lin2019auto}, MEGAN~\cite{sacha2021molecule}, Chemformer~\cite{irwin_chemformer_nodate}, GTA~\cite{seo_gta_2021}, and DualTF~\cite{sun_towards_nodate}.
%\end{itemize}

\subsection{Main Results}

\begin{table}[!b]
\begin{tabular}{@{}l@{}cr@{}ccccc@{}}
\toprule
\textbf{Method} &\textbf{TPL.}         & \textbf{K =} & \textbf{1} & \textbf{3}    & \textbf{5}    & \textbf{10}   & \textbf{50}   \\ \midrule
retrosim     & \CIRCLE  & & 37.3  & 54.7 & 63.3 & 74.1 & 85.3 \\
neuralsym    & \CIRCLE  & & 44.4  & 65.3 & 72.4 & 78.9 & 83.1 \\
MEGAN        & \Circle  & & 48.1  & 70.7 & 78.4 & 86.1 & 93.2 \\
G2Gs         & \LEFTcircle  & & 48.9  & 67.6 & 72.5 & 75.5 & -    \\
RetroXpert   & \LEFTcircle  & & 50.4  & 61.1 & 62.3 & 63.4 & 64.0 \\
GTA          & \Circle  & & 51.1  & 67.6 & 67.8 & 81.6 & -    \\
Hopfield     & \CIRCLE  & & 51.8  & 74.6 & 81.2 & 88.1 & 94.0 \\
GLN          & \CIRCLE  & & 52.5  & 69,0 & 75.6 & 83.7 & 92.4 \\
LocalRetro   & \CIRCLE  & & 53.4  & 77.5 & 85.9 & 92.4 & 97.7 \\ 
Dual-TF      & \Circle  & & 53.6  & 70.7 & 74.6 & 77.0 & -    \\
GraphRetro   & \LEFTcircle  & & 53.7  & 68.3 & 72.2 & 75.5 & -    \\
Chemformer   & \Circle & &54.3 & - & 62.3 & 63.0 & - \\
\midrule
\textbf{RetroKNN} & \CIRCLE & & \textbf{57.2}  & \textbf{78.9} & \textbf{86.4} & \textbf{92.7} & \textbf{98.1} \\ \bottomrule
\end{tabular}
\caption{Top-K exact match accuracy on the USPTO-50K dataset when the reaction type is unknown. The \CIRCLE, \LEFTcircle, and \Circle{} denote template-based, semi-template, and template-free, respectively. Systems are ordered by top-1 accuracy.}
\label{tab:uspto50k_unknown}
\end{table}

\begin{table}[!htbp]
\begin{tabular}{@{}l@{}cr@{}ccccc@{}}
\toprule
\textbf{Method} &\textbf{TPL.}          & \textbf{K =} & \textbf{1} & \textbf{3}    & \textbf{5}    & \textbf{10}   & \textbf{50}   \\ \midrule
retrosim      & \CIRCLE  & & 52.9  & 73.8 & 81.2 & 88.1 & -    \\
neuralsym     & \CIRCLE  & & 55.3  & 76.0 & 81.4 & 85.1 & -    \\
MEGAN         & \Circle  & & 60.7  & 82.0 & 87.5 & 91.6 & 95.3 \\
G2Gs          & \LEFTcircle  & & 61.0  & 81.3 & 86.0 & 88.7 & -    \\
RetroXpert    & \LEFTcircle   & & 62.1  & 75.8 & 78.5 & 80.9 & -    \\
GraphRetro    & \LEFTcircle  & & 63.9  & 81.5 & 85.2 & 88.1 & -    \\
LocalRetro    & \CIRCLE  & & 63.9  & 86.8 & 92.4 & 96.3 & 97.9 \\ 
GLN           & \CIRCLE  & & 64.2  & 79.1 & 85.2 & 90.0 & 93.2 \\
Dual-TF       & \Circle  & & 65.7  & 81.9 & 84.7 & 85.9 & -    \\
\midrule
\textbf{RetroKNN} & \CIRCLE & & \textbf{66.7}  & \textbf{88.2} & \textbf{93.6} & \textbf{96.6} & \textbf{98.4} \\ \bottomrule
\end{tabular}
\caption{Top-K exact match accuracy on the USPTO-50K dataset when the reaction type is given. The \CIRCLE, \LEFTcircle, and \Circle{} denote template-based, semi-template, and template-free, respectively. Systems are ordered by top-1 accuracy.}
\label{tab:uspto50k_known}
\end{table}

\begin{table}[!htbp]
\begin{tabular}{@{}l@{}cr@{}ccccc@{}}
\toprule
\textbf{Method} &\textbf{TPL.}         & \textbf{K =} & \textbf{1} & \textbf{3}    & \textbf{5}    & \textbf{10}   & \textbf{50}   \\ \midrule
Seq2Seq       &\Circle  & & 46.9  & 61.6 & 66.3 & 70.8 & -    \\
neuralsym     &\CIRCLE  & & 47.8  & 67.9 & 74.1 & 80.2 & -    \\
Transformer  &\Circle  & & 54.1  & 71.8 & 76.9 & 81.8 & -    \\
LocalRetro    &\CIRCLE  & & 54.1  & 73.7 & 79.4 & 84.4 & 90.4 \\
\midrule
\textbf{RetroKNN} &\CIRCLE & & \textbf{60.6}  & \textbf{77.1} & \textbf{82.3} & \textbf{87.3} & \textbf{92.9} \\ \bottomrule
\end{tabular}
\caption{Top-K exact match accuracy on the USPTO-MIT dataset. The \CIRCLE{} and \Circle{} denote template-based and template-free methods. Systems are ordered by top-1 accuracy.}
\label{tab:uspto_mit}
\end{table}

The experimental results of the USPTO-50K benchmark are shown in Table~\ref{tab:uspto50k_unknown} when the reaction type is unknown and in Table~\ref{tab:uspto50k_known} when the reaction type is given.
Meanwhile, the results on the USPTO-MIT benchmark are in Table~\ref{tab:uspto_mit}.
In these tables, we sort all systems by their top-1 accuracy and mark their type by filling the cycle symbols.
Our method (RetroKNN) is in the last row and highlighted in bold.

Comparing these accuracy numbers, we can find that our method outperforms the baseline systems with a large margin.
When the reaction type is unknown, we achieved 57.2 points top-1 accuracy and improved the backbone result from LocalRetro by 3.8 points, which is a 7.1\% relative gain.
When the reaction type is given, we also improve the top-1 accuracy by 2.8 points from 63.9 to 66.7.
Meanwhile, on USPTO-MIT, our method shows 60.6 points top-1 accuracy with a 6.5 points improvement or 12\% relative gain.
More importantly, these top-1 accuracies are also better than other strong baselines and state-of-the-art, demonstrating the effectiveness of our method.

At the same time, we achieved 78.9 points top-3 accuracy and 86.4 points accuracy in USPTO-50K when the reaction type is unknown, 
which are also much higher than baselines.
For the top-10 and top-50 accuracy, we get 92.7 and 98.1 points accuracy.
Considering that the accuracy is already very high, the improvement is still significant. 

To sum up, the local template retrial method efficiently improves the retrosynthesis prediction accuracy.

\section{Study and Analysis}
\subsection{Case Study}
\label{sec:case}
\begin{figure*}[!tb]
    \centering
    \includegraphics[width=\linewidth]{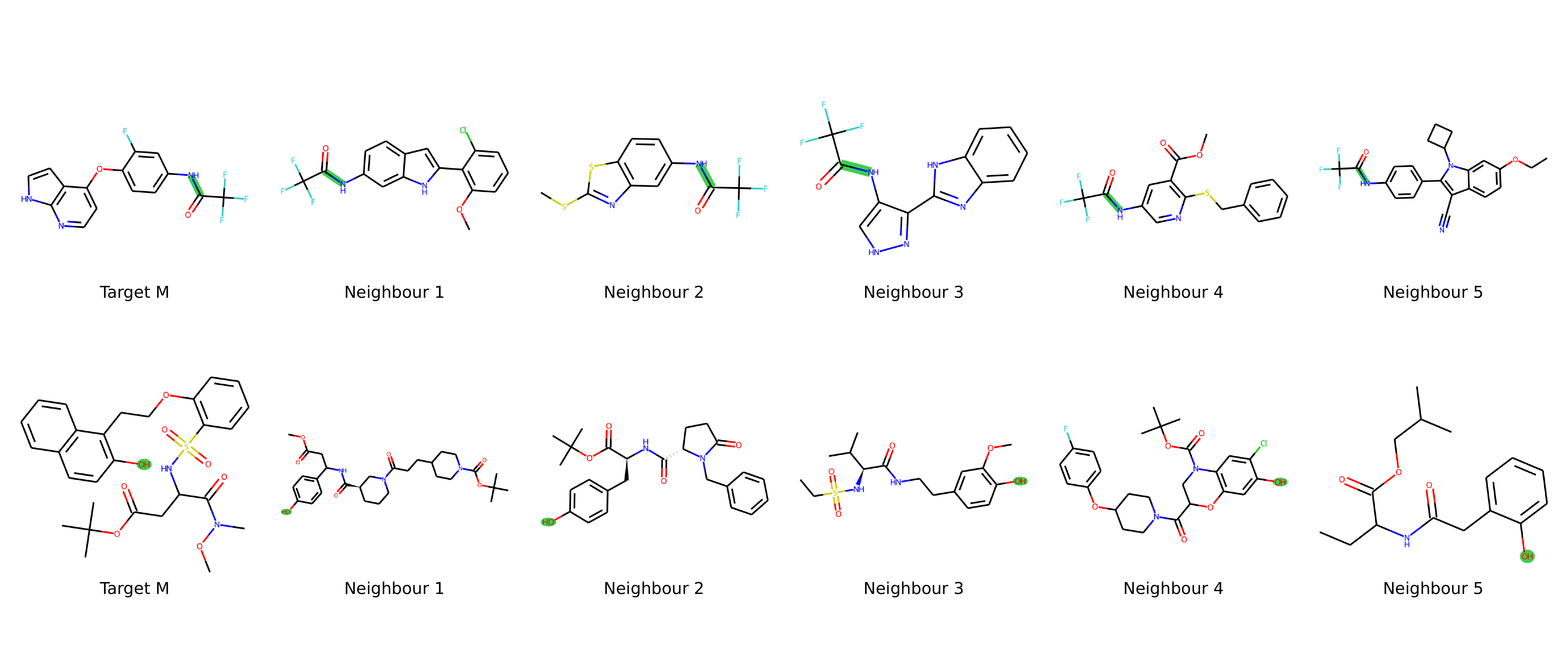}
    \caption{Case study of retrieved molecules. The bonds and atoms used in retrieval are highlighted by green background. The first column shows the target molecules, and the rest show five neighbourhood targets from the training data.}
    \label{fig:case}
\end{figure*}

\begin{table*}[!ht]
    \centering
    \begin{tabular}{@{}l|c|cc|c|ccccc@{}}
    \toprule
    \textbf{Target  Molecule}                & \textbf{GT.}   & $ \pmb{\lambda}$ & \textbf{T}   & \textbf{GNN} & \textbf{N1}           & \textbf{N2}           & \textbf{N3}          & \textbf{N4}          & \textbf{N5}  \\ \midrule
    \multirow{2}{*}{Cc1ccc(-c2cccnc2C\#N)cc1} &  \multirow{2}{*}{\textbf{b542}} & \multirow{2}{*}{0.96}    & \multirow{2}{*}{21.42} & \multirow{2}{*}{\textbf{b542}}      & b519  & b519 & b519 & b519 & b0  \\
                                              &      &         &       &           &(67.51)&  (77.35) & (77.35) & (77.35) & (104.00) \\ \midrule
    \multirow{2}{*}{\parbox{4.56cm}{CCOc1ccc(C[C@H](NC(=O)C( F)(F)F)C(=O)O)cc1}} & \multirow{2}{*}{\textbf{b524}} & \multirow{2}{*}{0.14} & \multirow{2}{*}{7.89} &  \multirow{2}{*}{b495} & \textbf{b524}  & b523  & b495& b495& b495 \\
                                                               &      &      &      &       & (22.79) & (33.84) &  (67.3) & (76.21) & (76.55) \\ \midrule
    \multirow{2}{*}{\parbox{4.59cm}{CC1(C)CC(=O)N(Cc2ccccc2)c2 ccc(C\#Cc3ccc(C(=O)O)cc3)cc21}} & \multirow{2}{*}{\textbf{a121}} & \multirow{2}{*}{0.02} & \multirow{2}{*}{19.36} & \multirow{2}{*}{a124} & \textbf{a121}& \textbf{a121}& \textbf{a121} &   a0 &   a0 \\
                                                                                            &      &      &       &      & (34.41) & (57.3) & (58.4) & (59.91) & (61.17) \\
    \bottomrule
    \end{tabular}
    \caption{Case study of parameter T and $\lambda$. The GT. denotes ground truth template id, GNN denotes the GNN prediction, and N1 to N5 denotes five neighbors. The prefix a, b of template id means it is an atom or bond template. We show each neighbor's distance in the brackets below template id. The correct predictions are highlighted in bold.}
    \label{tab:case_param}
\end{table*}

\noindent\textbf{Retrieval case study.}
To better understand if we can retrieve useful reactions by the hidden representations, we conducted case studies on the USPTO-50K datasets, and the results are shown in Figure~\ref{fig:case}.
We fist select an atom-template reaction and the first bond-template reaction from the data.
Next, we query the atom and bond store by the corresponding atom and bond.
Finally, for each retrieved template, we show the original target molecule in the training data, where the reaction atom/bond is highlighted by green background. 
The bond-template and atom-template reactions are available in the figure's first and second rows.
In each row, we first show the target molecule M of the reaction and then five neighbors of M.
From these cases, we can find that the neighborhoods retrieved by hidden representations can effetely capture the local structure of molecules.
For example, the carbon-nitrogen bond retrieves all neighbors in the edge-template reaction.
Moreover, all carbon atoms are surrounded by oxygen in a double bond~(\texttt{=O}) and a trifluorocarbon~(\texttt{-CF3}), and all nitrogen atoms are connected to an aromatic ring.
Meanwhile, for the node-template reaction, all retrieved atoms are the oxygen atoms that are connected to a phenyl.
In conclusion, retrieving molecules with hidden representations is efficient because it can capture the local structure well.
Therefore, we can improve the prediction accuracy by using the retrieved templates.

\noindent\textbf{Adapter case study.}
We show three representative cases for the effect of adapter in Table~\ref{tab:case_param}.
In each row, we show the target molecule and ground truth template id, then the $\lambda$ and T output by the adapter, and finally the GNN prediction and KNN retrieved neighbors.
When the GNN prediction is accurate in the first row, the adapter will generate a high $\lambda$ value (e.g., 0.96) so that the GNN output has a higher weight. 
However, when that is not the case (the second and third row), the $\lambda$ tends to be lower (e.g., 0.14), which gives more weight to KNN prediction.
Meanwhile, when only the N1 has the correct prediction (the second row), the adapter tends to output a small T (e.g., 7.89) to make the sharp distribution that gives more weight to N1's prediction.
On the contrary (the third row), the adapter tends to output a larger value (e.g., 19.36) so that more neighbors can contribute to the final output.
Moreover, our statistics show that when $\lambda < 0.5$, the GNN and KNN accuracy are 46.9\% and 69.2\%, showing that KNN is complementary to GNN prediction.
%From these observations, we can see the importance of the adapter for better gathering the predictions.

\subsection{Zero-shot and Few-shot Study}

\begin{figure}[!htbp]
    \centering
    \includegraphics[width=\linewidth]{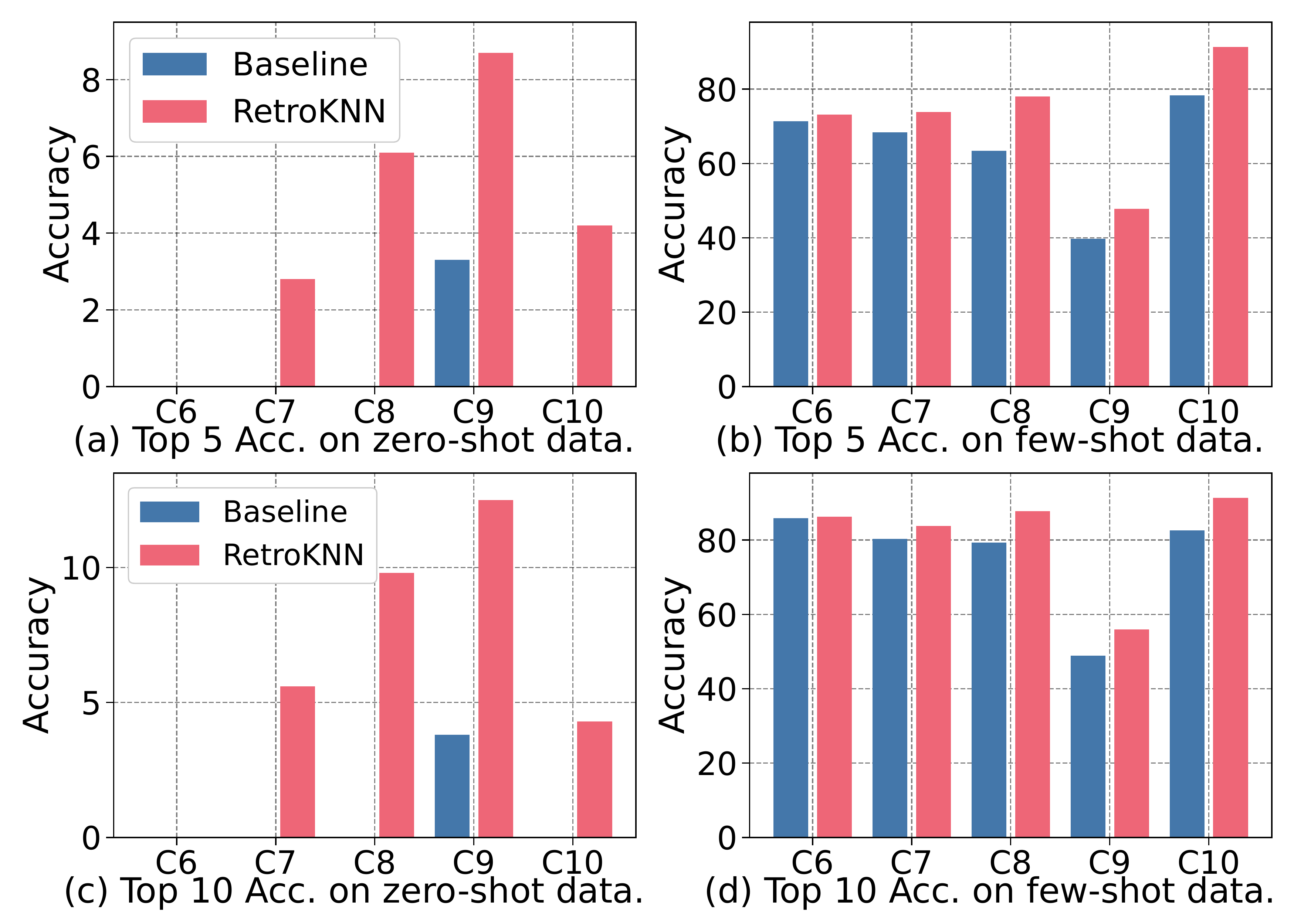}
    \caption{Top-5 (a, b) and top-10 (c, d) accuracy (Acc.) on the zero-shot (a, c) and few-shot (b, d) data. The columns C6 to C10 denote different reaction classes.}
    \label{fig:fs}
\end{figure}

We modify the USPTO-50K dataset to zero-shot and few-shot versions to study the domain adaptation ability of our method.
Specifically, in the USPTO-50K data, each reaction has its reaction class available in class 1 to 10.
To build the zero-shot data, we filter the train and validation data by removing all reactions with reaction class 6 to 10 and only keeping those with reaction class 1 to 5.
Similarly, to build the few-shot data, we only keep 10\% of reactions that have class 6 to 10.
Finally, we evaluate the performance of these new data with the LocalRetro baseline and our RetroKNN method.
The results are summarized in Figure~\ref{fig:fs}.
%In Figure~\ref{fig:fs}, we show the top-5 accuracy on the top half and top-10 accuracy on the bottom half.

From these plots, we notice that zero-shot is a challenging setting for conventional template-based methods, which is a known shortcoming of this kind of methods.
However, when combined with KNN, our system can generate meaningful results. 
For example, in reaction class 8, the RetroKNN haves 6.1 points top-5 accuracy and 9.8 points top-10 accuracy in the zero-shot data.
The few-shot setting is easier than the zero-shot because a few examples are available during training. Nevertheless, the RetroKNN also outperforms baseline on all reaction types. On average, the RetroKNN improved 8.56 points top-5 accuracy and 5.64 points top-10 accuracy. These results show that our method is can also improve the performance on zero/few-shot data, which are important scenarios in this field.

% % top-5 acc
% \begin{table}[!htbp]
% \begin{tabular}{lcccccc}
% \toprule
% \multirow{2}{*}{\textbf{Method}}   & \multirow{2}{*}{\textbf{Data}} & \multicolumn{5}{c}{\textbf{Reaction Type}} \\ \cmidrule(l){3-7} 
%                           &                       & \textbf{6}    & \textbf{7}   & \textbf{8}   & \textbf{9}   & \textbf{10}  \\ \midrule
% \multirow{2}{*}{Baseline} & ZS                    & 0    & 0   & 0   & 3.3 & 0   \\
%                           & FS                    & 71.4 & 68.4& 63.4&39.7 & 78.3\\ \midrule
% \multirow{2}{*}{RetroKNN} & ZS                    & 0    & 2.8 & 6.1 & 8.7 & 4.3 \\
%                           & FS                    & 73.1 & 73.8& 78.0& 47.8& 91.3\\ \midrule\midrule
% \multirow{2}{*}{Baseline} & ZS                    & 0    & 0   & 0   & 3.8 & 0   \\
%                           & FS                    & 85.9 & 80.3& 79.3& 48.9&82.6 \\ \midrule
% \multirow{2}{*}{RetroKNN} & ZS                    & 0    & 5.6 & 9.8 & 12.5 & 4.3 \\
%                           & FS                    & 86.3 & 83.8& 87.8& 56.0& 91.3\\                         
% \bottomrule
% \end{tabular}
% \caption{Top-5 (top) and top-10 (bottom) accuracy on the zero-shot (ZS) and few-shot (FS)  data.}
% \label{tab:fs}
% \end{table}

\subsection{Ablation Study}
\newcommand{\cycletext}[1]{\raisebox{.5pt}{\textcircled{\raisebox{-.9pt} {#1}}}}

We conducted an ablation study on the USPTO-50K dataset to study the contributions of different components, and the results are shown in Table~\ref{tab:ablaion}.
We show the top-1 accuracy in the table by comparing different systems.
The system~\cycletext{1} is the LocalRetro baseline without using KNN, which achieved 53.4 points accuracy.
In system~\cycletext{2}, we add the KNN without using the adapter.
To find the optimal paramters, we conduct comprehensive grid search on by $T \in \{1, 5, 25, 50\}$ and $\lambda \in \{0.1, 0.3, 0.5, 0.7, 0.9\}$, which leads to total 20 combinations.
We select the parameters by the validation loss and finally get the 56.3 points accuracy.
Furthermore, in system~\cycletext{3}, we add the adapter only for T and keep the $\lambda$ same as system~\cycletext{2}.
Similarly, we only add the adapter only for $\lambda$ in system~\cycletext{4}.
The system~\cycletext{5} is the full RetroKNN model.

Comparing the system~\cycletext{1} with others that using KNN, we can find that introducing KNN to this task can effectively improve the model performance.
These numbers show that the local template retrieval is vital for the system.
Meanwhile, comparing system~\cycletext{3}\cycletext{4} to system~\cycletext{2}, we notice that adding both T and $\lambda$ adapter is helpful.
Finally, when both parameters are adaptively predicted in system~\cycletext{5}, the accuracy can be boosted to 57.2, showing that they can work together effectively.
Therefore, all components are necessary for this system.

\begin{table}[!tb]
    \centering
    \begin{tabular}{llc}
    \toprule
        \textbf{ID} & \textbf{System}   &  \textbf{Accuracy} \\
        \midrule
        \cycletext{1} & Baseline &  53.4 \\
        \cycletext{2} &~~~~ + KNN & 56.3 \\
        \cycletext{3} &~~~~ + KNN, adaptive T & 56.7 \\
        \cycletext{4} &~~~~ + KNN, adaptive $\lambda$ & 56.8 \\
        \cycletext{5} &~~~~ + KNN, adaptive T, adaptive $\lambda$ & 57.2 \\
    \bottomrule
    \end{tabular}
    \caption{Ablation study on the USPTO-50K dataset when the reaction type is unknown.}
    \label{tab:ablaion}
\end{table}

\subsection{Retrieved Templates Size}

%In this section, we study the impact of the KNN parameters.
In Table~\ref{tab:para_k}, we show how the number of retrieved reactions (i.e., K of KNN) affects the model performance.
More specifically, in the KNN search, we set the K $\in [1, 4, 8, 16, 32] $, then train adapters for each of them.
Finally, we report the top-1 accuracy in the table.

From these results, we first observe that only adding one retrieved template (K=1) can improve the accuracy from 53.4 to 55.6.
When K is $\ge$ than 4, the accuracy can be further improved to around 57 points.
There will be no further significant improvement when more reactions are retrieved, nor will more received templates hurt the performance.
We suppose it is because there is already enough information to improve the accuracy as the templates far from the query will contribute less to the prediction.

% Second, we show the histogram of parameter T and $\lambda$ in Figure~\ref{fig:para_dist}.
% We conduct inference on the USPTO-50 validation set to get these numbers by collecting all output of the adapter.
% We can find that instead of finding an optimal global value, the adapter can adaptively predict different values for different targets.
% For the parameter T, sometimes it prefers a low value (e.g., less than 20), and sometimes it uses a higher value.
% Meanwhile, for the parameter $\lambda$, although often the values around 0.9 are used, it will also use a lower value when the adapter thinks the KNN prediction is more accurate.
% These results can help explain the effectiveness of the adapter network.

\begin{table}[!tb]
    \centering
    \begin{tabular}{@{}c|ccccc@{}}
    \toprule
    \textbf{\#Retrieved reactions} & 1 & 4 & 8 & 16 & 32 \\
    \midrule
    \textbf{Accuracy}  &  55.6 & 57.4   & 57.1  &    56.9 & 57.2 \\
    \bottomrule
    \end{tabular}
    \caption{Study on the number of retrieved reactions by KNN.}
    \label{tab:para_k}
\end{table}

% \begin{figure}[!htbp]
%     \centering
%     \includegraphics[width=\linewidth]{param.pdf}
%     \caption{Distribution of parameter T and $\lambda$ on the USPTO-50 validation set.}
%     \label{fig:para_dist}
% \end{figure}

\subsection{Inference Latency}

In Table~\ref{tab:latency}, we study the datastore size and the inference latency.
The last two rows present the latency with or without retrieval during inference, which are measured on a machine with a single NVIDIA A100 GPU.
Each latency value, which is the average run time per reaction, is measured with ten independent runs.
%We use batch size 16 and measure the latency with ten independent runs.
%Each number in the table is the average run time per reaction.
In the USPTO-50K dataset, we observe that the average latency increased from 2.71 ms to 3.31 ms, which is about 0.6 ms for each reaction. 
The extra latency is a little more prominent for the USPTO-MIT dataset because it is about ten times larger than the USPTO-50K. 
However, considering the hours or even days that a more accurate system can save for chemists, the extra ten-millisecond cost is not a real obstacle to the practical use of this method.
Finally, some work~\cite{he2021efficient,meng2021fast} show that the KNN speed can be further accelerated, and we would like to add these techniques in future work.

\begin{table}[!tb]
    \centering
    \begin{tabular}{c|cc}
    \toprule
    \textbf{Dataset} & \textbf{USPTO-50K} & \textbf{USPTO-MIT} \\
    \midrule
    $|\set{D}_{\text{train}}|$ & 40k & 409k \\
    $|\set{S}_A|$ & 1,039k & 10,012k \\
    $|\set{S}_B|$ & 2,241k &21,495k \\
    Latency w/o KNN & 2.71 $\pm$ 0.02 ms & 3.51 $\pm$ 0.05 ms \\
    Latency w/ KNN & 3.31 $\pm$ 0.09 ms & 14.69 $\pm$ 0.29 ms \\
    \bottomrule
    \end{tabular}
    \caption{Study of the datastore size and inference latency.}
    \label{tab:latency}
\end{table}

\section{Related Work}

\subsection{Retrosynthesis Prediction}

% %Retrosynthesis prediction is an essential task for scientific discovery.
% The ML assisted methods can be mainly categorized into two types: template-free methods and template-based methods.
% A classical template-free method will predict the reactant SMILES directly~\cite{liu_retrosynthetic_2017,tetko_state---art_2020,irwin_chemformer_nodate,wan_retroformer_2022}.
% On the contrary, a template-based method~\cite{coley_computer-assisted_2017,segler_neural-symbolic_2017, dai_retrosynthesis_2020,seidl_modern_2021,chen_deep_2021} will first predicts reaction templates.

Retrosynthesis prediction is an essential task for scientific discovery and have achieved promising results in recent years~\cite{segler_neural-symbolic_2017,liu_retrosynthetic_2017,coley_computer-assisted_2017,tetko_state---art_2020,irwin_chemformer_nodate,dai_retrosynthesis_2020,yan2020retroxpert,seidl_modern_2021,chen_deep_2021,shi_graph_2021,somnath_learning_2021,wan_retroformer_2022}.
%Moreover, some recent work~\cite{shi_graph_2021,yan2020retroxpert,somnath_learning_2021} use a semi-template method for this task.
A few research also use retrieval mechanisms for this task.
For example,~\citet{seidl_modern_2021} use Hopfield networks to select templates, and~\citet{lee_retcl_2021} use retrieval method to fetch molecules from a database.
Being differently, we are the first to combine deep learning and KNN retrieval in this task.

% For example,~\citet{seidl_modern_2021} use Hopfield networks to select templates and  
% The key difference between this work and ours is that they only use templates as store.
% Meanwhile,~\citet{lee_retcl_2021} use retrieval method to fetch molecules from a database as reactants.
%Being differently, we retrieve the reaction templates rather than molecules. Therefore, our results are more diverse and explainable.

\subsection{Retrieval Methods }
Retrieving from data store or memory to improve the machine learning model's performance is an important research topic.
SVM-KNN~\cite{zhang2006svm} first combines the SVM and KNN for recognition tasks.
%More recently,~\citet{tu2018learning} build a continues cache for translation task and \citet{wu2020taking} use a note dictionary for pretrainig.
Furthermore, the KNN-LM~\cite{khandelwal20generalization} and KNN-MT~\cite{khandelwal2021nearest} have shown promising results when combining KNN with Transformer networks.
Meanwhile, \citet{he2021efficient,meng2021fast} study the speed of retrival methods and~\citet{zhen2021ada} study the adaptation problem.
However, we are the first to combine the strong capability of KNN with GNN and use them on the retrosynthesis task.

\section{Conclusion}
Retrosynthesis prediction is essential for scientific discovery, especially drug discovery and healthcare. In this work, we propose a novel method to improve prediction accuracy using local template retrieval. We first build the atom and bond stores with the training data and a trained GNN and retrieve templates from these stores during inference. The retrieved templates are combined with the original GNN predictions to make the final output. We further leverage a lightweight adapter to adaptively predict the weights to integrate the GNN predictions and retrieved templates. We greatly advanced the prediction performance on two widely used benchmarks, the USPTO-50K and USPTO-MIT, reaching 57.2 and 60.6 points for top-1 accuracy. These results demonstrate the effectiveness of our methods.

 \section*{Acknowledgements}
 We would like to thank the anonymous reviewers for their insightful comments.
This work was supported by National Natural Science Foundation of China (NSFC Grant No.~62122089 and No.~61876196), Beijing Outstanding Young Scientist Program NO. BJJWZYJH012019100020098, and Intelligent Social Governance Platform, Major Innovation \& Planning Interdisciplinary Platform for the ``Double-First Class'' Initiative, Renmin University of China.
We also wish to acknowledge the support provided and contribution made by Public Policy and Decision-making Research Lab of RUC.
Rui Yan is supported by Beijing Academy of Artificial Intelligence (BAAI).

\bibliography{aaai23}

\end{document}